# Finite element model selection using Particle Swarm Optimization

Linda Mthembu[1], Tshilidzi Marwala[2], Michael I. Friswell[3], Sondipon Adhikari[4]

[1]Visiting Researcher, Department of Electronic and Computer Engineering, Faculty of Engineering and Built Environment, University of Johannesburg, PO Box 17011, Doornfontein 2028, South Africa.

[2]Executive Dean Faculty of Engineering and Built Environment, University of Johannesburg, PO Box 17011, Doornfontein, 2028, South Africa, South Africa.

[3]Professor of Aerospace Structures, School of Engineering, Swansea University, Singleton Park, Swansea SA2 8PP, United Kingdom.

[4]Chair of Aerospace Engineering, School of Engineering, Swansea University, Singleton Park, Swansea SA2 8PP, United Kingdom

## Abstract

This paper proposes the application of particle swarm optimization (PSO) to the problem of finite element model (FEM) *selection*. This problem arises when a choice of the best model for a system has to be made from set of competing models, each developed a priori from engineering judgment. PSO is a population-based stochastic search algorithm inspired by the behaviour of biological entities in nature when they are foraging for resources. Each potentially correct model is represented as a particle that exhibits both individualistic and group behaviour. Each particle moves within the model search space looking for the best solution by updating the parameters values that define it. The most important step in the particle swarm algorithm is the method of representing models which should take into account the number, location and variables of parameters to be updated. One example structural system is used to show the applicability of PSO in finding an optimal FEM. An optimal model is defined as the model that has the least number of updated parameters and has the smallest parameter variable variation from the mean material properties. Two different objective functions are used to compare performance of the PSO algorithm.

**Nomenclature**

| | |
|---|---|
| FEM | Finite element model. |
| FEMU | Finite element model updating. |
| PSO | Particle Swarm Optimization. |
| $d$ | Model Dimension. |
| $m_{id}$ | i-th Finite element model position at parameter d. |
| $v_{id}$ | i-th Finite element model velocity at parameter d. |
| $p_{id}$ | Best position for the ith Finite element model at parameter d. |
| $p_{gd}$ | Global best finite element model at parameter d. |
| $w_k$ | The k-th dimension inertia weight. |
| $M_{max}, M_{min}$ | Maximum and minimum model position respectively. |
| $V_{max}, V_{min}$ | Maximum and minimum model velocity respectively. |
| $E_i$ | Young's Modulus. |
| AIC | Akaike Information Criterion. |
| SSE | Sum of Squared Errors. |
| $\mu$ | Mean |
| $msrd_{data}$ | Measured structural results. |
| $fem_{results}$ | Finite element model results. |

## 1. Introduction

The finite element model updating (FEMU) problem arises due to the mismatch between the initial finite element model results and the measure real system results [7, 9]. The modeller's problem is then to determine which aspects or feature of the initial model are uncertain or incorrectly modelled. This is effectively a system-identification problem [15]. In classic system-identification the real measured system/data is approximated by a set of mathematical equations, usually polynomial equations whose parameters or order is unknown and is to be identified. In this paper the system is a structural system which is described by a finite element model and what needs to be identified are the uncertain parameters of the initial FE model [8, 9].

In the literature there are two main directions to finite element model updating (FEMU); direct and indirect (iterative) methods [9]. In the direct model updating paradigm [4, 9] the measured modal data are directly equated to the model modal values thus freeing up the system matrices (the mass, damping and or stiffness matrix elements) for updating. This approach often results in unrealistic system matrix element magnitudes, for example, physically unrealisable mass elements. In the indirect or iterative model updating approach the updating problem is formulated as an optimization problem, often approached by the use of least squares, maximum likelihood, eigenvalue sensitivity, genetic algorithm optimization and Bayesian approaches [1, 2, 8, 9, 12, 13, 16, 17, 18, 20, 21 and 26]. This paper proposes an iterative scheme to model updating.

This paper continues addressing the FEMU problem with the approach proposed in [20]. This approach is based on directly answering three standard questions in FEM updating; (a) which aspects of the models do we need to update? (b) how are we to update these models? (c) is the updated model the best one? In [20, 21] a particular Bayesian approach to model updating was presented. The methods considered updating FE models using Bayesian stochastic techniques from which one can determine the best model in a given group. The approaches presented in [20, 21] and in this paper effectively propose performing finite element model updating within the model selection framework. This whole approach is based on the premise that different analysts would differ (for example the updating parameters of the GARTEUR structure in [14]) on which aspects of the initial finite element model are incorrectly modelled to answer question (a), and how to proceed with the model updating (when answering question (b)). This disagreement between researchers on the first two questions often makes it difficult to compare FEM updating results of the same structure let alone different updating techniques. To answer questions (b) and (c) we assume that a number of pre-existing potentially correct finite element models of a *particular* structure have been developed. These models could have been generated through engineering judgement or competing analysis by different people/techniques. Given the problem described in the previous paragraph, the problem is then to develop a method of both updating all the models and then selecting the best performing one. Ideally one would like an all-in-one procedure of both updating and selection. This would make it easier to compare the model updating results as they would have been updated 'similarly'. The particle swarm optimization framework proposed in this article allows this simultaneous updating of all competing models and the identification of the best model in the given group.

## 2. Particle Swarm optimization(PS0)

Particle swarm optimization was first developed by [10]. PSO is a population-based stochastic search algorithm inspired by the social-psychological behaviour of biological entities in nature when they are foraging for resources. The population/swarm of entities in nature could be that of birds, fish and or ants etc searching for food [5, 10, 11]. Each entity in the swarm is able to dynamically adapt individually and through group influence to the environment while in search of resources. The swarm adapts by stochastic moving towards previously good regions in the environment. This means the movement of the swarm in the search space has some random elements to it but this movement tends to converge to optimal points in the search space.

The swarm behaviour metaphor has been adopted by the evolutionary computation community [5, 11, 13] where the biological entities are called particles, the swarm is called a population, the environment is the solution space and the resource is the solution to the problem. One of the main differences between evolutionary and classic swarm based algorithms is the way the particles interact. In a typical evolutionary algorithm (e.g. Genetic algorithm [13]), particles combine and mutate within a population and over generations. In swarm based approaches for example, particles communicate instead of merging. There is no evidence of one method being superior to the other but consensus is that these methods are well suited to problems where the solution search space is too large to search exhaustively [6, 11, 24, 26].The practical interpretation of this analogy is that a number of particles stochastically 'move' through the problem search space searching for the minimum/maximum solution point. This means *each* considered particle can potentially find the optimal solution to the problem. In the FEMU context and approach proposed in this paper; each particle is a potentially correct model to the finite element modelling problem. The particle search space is defined by the number of updating parameters in the models; specifically the maximum number of updating parameters defines the complete search space. Obviously if the models do not have the same number of free parameters then some models will only search a subset of the full search space nonetheless they will be embedded in the full space (see section 4.1 for the chosen particle representation).

In the next section the mathematical operators of the PSO algorithm are presented.

2.1 PSO Operators

In the PSO algorithm each particle is described by its vector position in the search space as in equation 1 below;

$$m_i = \{m_{i1}, m_{i2}, m_{i3}....m_{id}\} \quad (1)$$

where *i* is an arbitrary particle and *d* is the problem dimension defined by the maximum number of updating parameters for the models. The particle position features ($m_{id}$'s) are the potential solution variables. This means to evaluate each particle on the problem one substitutes the particle position to the model/function. The rate of the i-th particle's position change, the velocity, is represented by;

$$v_i = \{v_{i1}, v_{i2}, v_{i3}....v_{id}\} \quad (2)$$

Each particle also stores its (local) best ever position as it searches the problem space. This is represented by;

$$p_i = \{p_{i1}, p_{i2}, p_{i3}....p_{id}\} \quad (3)$$

The swarm also has a record of the best ever position by any particle, this is known as the global best solution (socially) which is represented by;

$$p_g = \{p_{g1}, p_{g2}, p_{g3}....p_{gd}\} \quad (4)$$

On each iteration of the PSO algorithm the position and velocity of each particle are updated. The particle velocity is updated through the following equation;

$$v_{ik}(t) = w_k v_k(t-1) + c_1 r_1 (p_{ik} - m_{ik}) + c_2 r_2 (p_{gk} - m_{ik}) \quad (5)$$

and then the particle position is updated using equation 6;

$$m_{ik}(t) = m_{ik}(t-1) + v_{ik}(t) \quad (6)$$

where $\{i \in 1...m\}$ and $\{k \in 1...d\}$ which means each particle's position and velocity parameters /dimensions are updated on each iteration of the algorithm. See section 4 for the PSO algorithm pseudo code. In equation 6, $c_1$ and $c_2 \in \mathbb{R}$ are constants weighting that normally vary between 2 and 4 [10]. The random constants $r_1, r_2 \sim U[0,1]$ introduce randomness to the search process. In order to prevent the tendency of the particle position and velocity to explode in magnitude, $M_{max}$, $M_{min}$, $V_{max}$ and $V_{min}$ are defined for each particle dimension. Thus if

$$m_{ik} > M_{max} \text{ then } m_{ik} = M_{max}$$
$$m_{ik} < M_{min} \text{ then } m_{ik} = M_{min}$$

Similarly if

$$v_{ik} > V_{max} \text{ then } v_{ik} = V_{max}$$
$$v_{ik} < V_{min} \text{ then } v_{ik} = V_{min};$$

The setting of these limits on each dimension of the problem would depend on the analyst's understanding of the problem. The types of constraints are also very much dependent on the problem, for example some particle dimensions might be known to be constrained to be positive values, and in that case the absolute $|M_{max}|$ or $|V_{max}|$ might be applicable [10,11, 25].

A modification to the original PSO algorithm by [24] introduced $w_k$, the inertia weight variable. It controls the influence of the previous velocity on the current velocity value. An adaptive inertia weight is often used to improve the algorithm's search from an initially explorative (global) search to a more local search as this variable decreases. This also has a tendency to improve the algorithm's convergence rate [11, 24]. This variable is specified by the starting weight ($w_{start}$), $w_f$ is the fraction of iterations over which the inertia weight is decreased and $w_{end}$ the final inertia weight value The initial $w_k$ in equation 5 is $w_{start}$ and it is decrease by $w_k = w_k - w_{dec}$ where

$$w_{dec} = \frac{w_{start} - w_{end}}{N - w_f} \qquad (7)$$

from the first iteration up to iteration N x $w_f$, thereafter $w_k$ is $w_{end}$. The ($p_{ik}$- $m_{ik}$) term in equation 5 measures how far each particle is currently from its personal best position (local) and ($p_{gk}$- $m_{ik}$) measures how far each particle is from the global (social) best particle in the swarm. This means the middle term in equation 6 tends to control the particle's velocity based on the particles own best position while the last term allows the particle to be influenced by the best performing particle in the swarm.

In the next section we present the finite element models and propose a particular representation of these in the particle swarm context.

## 3. Finite Element Models

The finite element models updated in this paper were all developed from the unsymmetrical h-beam structure In [20] and previously used in [19]. The details of the modelled beam are described in the next section.

3.1 Unsymmetrical H-Beam

A simple unsymmetrical h-beam shown in figure 1 is modelled. This unsymmetrical h-beam is suspended on rubber bands (see [19] for more details on the structure and experimental set-up). The measured natural frequencies of interest of this structure occur at; 53.9Hz, 117.3Hz, 208.4Hz, 254Hz and 445Hz which correspond to modes 7, 8, 10, 11 and 13 respectively. The aluminium beam material has a Young's Modulus of $7.2 \times 10^{10}$ Pa, the beam length is 600mm with a width of 32.2mm and a section thickness of 9.8 mm. The left edge has a length of 400mm, the right edge length 200mm and a density of 2793 kg/m$^3$.

Figure 1 shows that the beam is divided into elements numbered from one to twelve. Each cross-sectional area is 9.8mm by 32.2 mm. Each finite element model used standard isotropic material properties and Euler Bernoulli beam elements to approximate the beam sections of the structure. The beam is free to move in all six degrees.

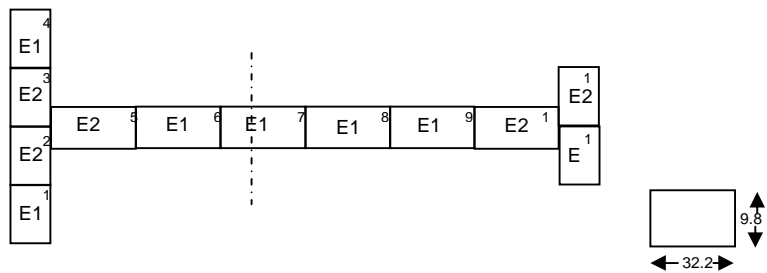

**Figure 1**: Model $M_2$ of a 12 Element Unsymmetrical H beam.

3.2 The mathematical models of the beam

All models in this example assume the only uncertain beam property is its Young's Modulus (E) value. To design different models of the beam, beam elements are grouped differently. The beam is modelled by eight competing models, $m_i$, $i = 1...8$. Model $m_1$ assumes the whole beam's Young's modulus is the updating parameter to be updated from the average given material value. Model $m_2$ has two parameter, $E_1$ and $E_2$; the elements numbered 1,4, 6,7,8,9 (all forming parameter $E_1$) are to be varied equally while elements 2, 3, 5, 10, 11, 12 (all $E_2$) are to be varied equally (see figure 1 for the element numberings).

Model $m_2$ models the elements connected near the structural joints as one parameter and those away from the joints as another. Model $m_8$ assumes the left edge together with the first horizontal element, the horizontal section and right edge together with the last horizontal element are best updated differently, thus the three parameter arrangement. Table 3 lists the rest of the models and their parameterizations.

| Model Identity | Number of model parameters | Parameter Labels | Element grouping |
|---|---|---|---|
| $m_1$ | 1 | $E_1$ | {1-12} |
| $m_2$ | 2 | $E_1$ & $E_2$ | {1,4,6-9} & { 2,3,5,10-12} |
| $m_3$ | 3 | $E_1$ $E_2$ $E_3$ | {1,4,6-9}, {2,3,11,12} & {5,10} |
| $m_4$ | 4 | $E_1$ $E_2$ $E_3$ $E_4$ | {1,4,6-9}, {2,3} {11,12} & {5,10} |
| $m_5$ | 5 | $E_1$ $E_2$ $E_3$ $E_4$ $E_5$ | {1,4,6-9}, {2,3} {11,12},{5} & {10} |
| $m_6$ | 2 | $E_1$ $E_2$ | {1,2,3,4} & {5-12} |
| $m_7$ | 2 | $E_1$ $E_2$ | {1- 6} & { 7-12} |
| $m_8$ | 3 | $E_1$ $E_2$ $E_3$ | {1-5}, {6-9} & (10,11,12} |

**Table 1**. Model Parameterization

Perhaps the most important step in the implementation of the PSO algorithm is the choice of particle representation. This fundamentally dictates the problem search space and the ease of algorithm implementation. The next section presents the model representation adopted in the current finite element model updating procedure.

### 4. PSO Algorithm

4.1 Particle Representation

Each particle or finite element model ($m_{i...m}$) is described by the following $E_i$ vector arrangement:

$m_1 = [E_1,0,0,0,0]$; $m_2 = [E_1,E_2,0,0,0]$; $m_3 = [E_1,E_2,E_3,0,0]$; $m_4 = [E_1,E_2,E_3,E_4,0]$;

$m_5 = [E_1,E_2,E_3,E_4,E_5]$; $m_6 = [E_1,E_2,0,0,0]$; $m_7 = [E_1,E_2,0,0,0]$; $m_8 = [E_1,E_2,E_3,0,0]$;

where in each case

$$E_{i...5} = \mu + q\sqrt{0.5e20}\text{N.m}^{-2}. \qquad (8)$$

and the mean $\mu = 7.2e10\text{N.m}^{-2}$.

The *q* variable samples random numbers from a normal distribution between [-∞, ∞].The parameter *location* of the models as described in Table 1 nullifies the concern that models $m_2$, $m_6$ and $m_7$ seem to be described by the same parameter vector.

This choice of model representation sets the problem search space to five dimensions. Even though all the models search the five dimensional space, each is actually constrained to only searching a particular manifold of the space. This contextually means each subspace is assumed to have an optimum somewhere which the particle is suppose to find guided by individualistic and social performance. This means if model $m_5$ finds the best solution within the group (thus it is $p_g$) at some coordinate/parameter values, all the other particles will adapt towards model $m_5$'s parameter values. In this particle representation, all the particle parameters will incrementally change towards model $m_5$'s values, even the zeros in the particle vector description. But since, for example, model $m_2$, $m_6$ and $m_7$ are only dependent on the first and second updating values, it does not matter what happens to the zero features! Each model will somewhat also resist moving towards model $m_5$'s coordinates by also incrementally moving towards their own previous best positions.

4.2 PSO Pseudo-Code

In this section the FEM-PSO pseudo-code algorithm together with the parameter settings used in the simulations is presented.

```
FEM-PSO Algorithm Pseudo Code
%Set constants
N = 500;         The max number of iterations
c₁ = c₂ = 2;    Individual and Group Influence
m = 8;           Number of potentially correct models
wstart = 1.2;   Initial inertia weight
wf = 0.5;        Inertia decrement factor
wend = 0.4;     Final inertia weight
%Initialise
w_dec   - Using equation 7.
m_i..m  - Randomly initialise the models
p_i..m  - Randomly initialise particle best solution (In this paper initial p_i=m_i)
% Compute
F(m_i..m) - Calc model Fitness using the objective function in section 4.4
Identify P_g
Iteration =1
Repeat
  While iteration< N do
        for all m_i..m do
          for each dimension d
              Calculate particle velocity(v_i) using equation 5
              Update particle position (m_i) using equation 6
           end for
           Compute F(m_i)
           Update p_i  if m_i(t) > p_i(t-1)
        end for
        Update p_g if any p_i > p_g(t-1)
        If iteration < |N w_f|
           w_k = w_k-w_dec
          end if
        iteration = iteration +1;
    end while
return p_g
```

**Table 2.** The FEM-PSO algorithm pseudo code

Obviously if the original PSO algorithm is implemented i.e. the inertia variable is eliminated, the last If statement condition is not executed and the velocity equation does not have $w_k$ on the first term. The next section describes the particle fitness functions used in all the simulations.

4.3 Model parameter constraints

In our FEMU problem the constraints placed on the particle velocity and position in the algorithm were as follows; the maximum parameter magnitude $M_{max}$ for each dimension was set at 7.5e10 N.m$^{-2}$ and the minimum, $M_{min}$, was set at 5.5e10. The maximum velocity magnitude was set to the difference between $M_{max}$ and $M_{min}$. The minimum velocity was set to 1e9 N.m$^{-2}$. This means the velocity in this algorithm was tracing a factor of the standard deviation of the parameter values (i.e. the second term in equation 8) and the particle position was determining the mean parameter value ($E_i$) in equation 8.

4.4 Objective Functions

A number of fitness or objective functions are available in the scientific literature. In the FEMU problem Occam's razor is very much applicable. This means one seeks a model with the fewest updating parameters that will produce FE model results closest to measured lab results. In this paper we compare two objective/fitness functions; the Akaike Information Criterion (AIC) [25] and the Squared Sum of Errors (SSE). The AIC function is given by the following equation:

$$AIC = n\log(\sigma^2) + 2d \tag{9}$$

where,

$$\sigma^2 = \frac{\sum_{i=1}^{n}(msrd_{data} - fem_{results})^2}{n}, \tag{10}$$

and $d$ is the number of model parameters, $n$ is the number of measured modes, $msrd_{data}$ is the measured data and $fem_{results}$ are the finite element model results. The squared sum of errors is given by

$$SSE = \frac{\sum_{i=1}^{n}(msrd_{data} - fem_{results})^2}{2} \tag{11}$$

As it can be seen from equations 9 to 11, the first term of equation 9 is effectively the SSE and is commonly referred to as the data-fit term [3, 20, 21, 23] and the second term is known as the model complexity penalty term. This assumes a models' complexity is determined by the number of free variables. This is not always the correct way to define model complexity as argued in [22, 23]. The minimization/maximization of the squared sum of error objective function does not account for model complexity but only measures how well a given model fits the data. We would expect the implementation of the AIC function as the objective function in the PSO algorithm to be biased to models with fewer parameters.

## 5. Simulation Results

The simulations presented in this section are all modelled using Version 6.0 of the Structural Dynamics Toolbox (SDT®) for Matlab. A number of simulations were run using different setting of the PSO algorithm parameters. Initially the number of iterations in all settings was set to $N$=1000 but it was found that the algorithm consistently converged before N= 500 iterations. In each of the experiments the convergence figures will only focus on the main convergence part of the graph, where necessary the figure will be expanded to show 500 iterations.

5.1 Simulation settings

| PSO Parameter | Simulation No.1 | Simulation No. 2 | Simulation No. 3 | Simulation No. 4 |
|---|---|---|---|---|
| $C_1$ Local influence | 2 | 2 | 2 | 2 |
| $C_2$ Global Influence | 2 | 2 | 2 | 2 |
| w | 0 | 0 | Adaptive | Adaptive |
| Objective function | AIC | SSE | AIC | SSE |

**Table 3**. PSO Simulation parameter settings

5.1.1 Simulation number 1

In this simulation the original PSO algorithm is implemented on the FEMU problem, this means there is no inertia variable as shown in table 3. Figure 2 shows the convergence plot of the AIC objective function over the 200 iterations of the algorithm. Figure 3 shows the variation of the best particle ($p_g$) in the swarm over 10 iterations.

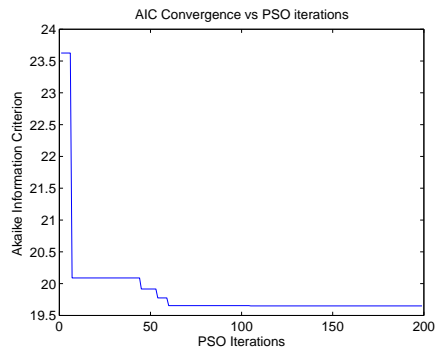

**Figure 2**. Akaike Information Criterion (AIC) convergence vs. PSO iterations.

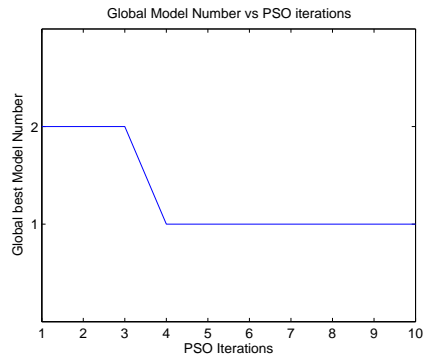

**Figure 3**. Global Model Number vs. PSO iterations

Figure 2 illustrates that the PSO algorithm rapidly converged close to the ultimate minimum error within the first 80 algorithm iterations. The global best model in this simulation started off as model $m_2$ but after 3 iterations it changed to model $m_1$ and remains unchanged for the rest of the simulation. The model order, based on the minimum of the objective function for these PSO settings was $m_1$, $m_6$, $m_2$, $m_7$, $m_8$, $m_3$, $m_4$ and then $m_5$.

5.1.2 Simulation number 2

This simulation is the same as simulation 1 except the objective function has been changed to SSE. Figure 4 shows the convergence plot of the SSE objective function over the first 100 algorithm iterations. Figure 5 illustrate the convergence behaviour of global best model over 100 iterations.

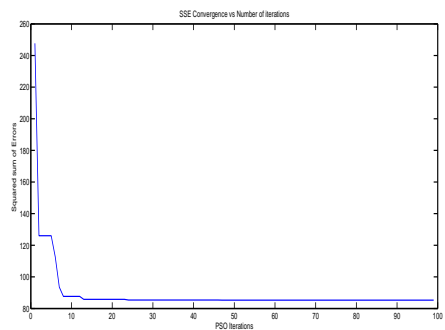

**Figure 4**. Squared sum of Errors (SSE) vs. PSO iterations

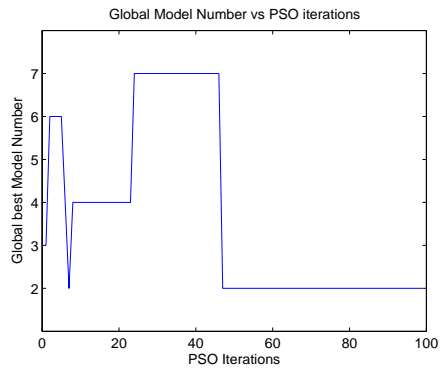

**Figure 5**. SSE Global best model for the first 100 PSO iterations.

The objective function (in Figure 4) did not improve much after 100 iterations even thou the global model changed. A relatively small improvement occurred (not shown) due to model $m_6$ becoming the global best model after 300 iterations. It is clear from figure 5 that the objective function had a significant role in the updating of the model parameters. Initially model $m_3$ was the $p_g$ then the global best changed to being $m_6$, $m_2$, $m_4$, $m_7$, $m_2$ then finally model $m_6$ was again the $p_g$. The final objective function based model order in this simulation was $m_6$, $m_1$, $m_2$, $m_5$, $m_4$, $m_8$, $m_3$ and then $m_7$. This is different to the results in simulation 1 where the less complex models attained lowest errors.

5.1.3 Simulation number 3

In simulation number 3 the PSO algorithm has been changed by introducing the adaptive inertia weight. As mentioned in section 2.1 the inertia parameter allows the algorithm to initially explore a wider search area and near the end to exploit the local search space.

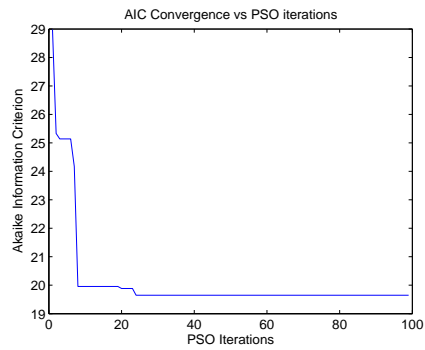

**Figure 6**. Akaike Information Criterion (AIC) convergence vs. PSO iterations

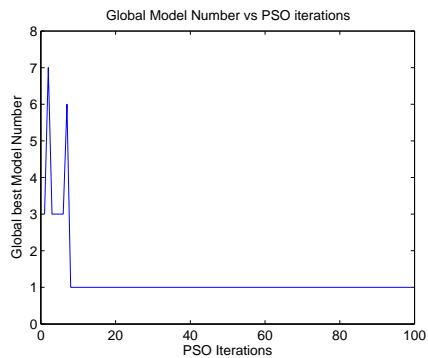

**Figure 7**. AIC Global model convergence vs. PSO iterations

This is evident from figure 6, where the search converged much quicker than in figure 2 whilst using the same objective function. Figure 7 also supports the initial explore to local exploitation concept because a number of models were initially $p_g$ as opposed to figure 3 but towards the end a firm favourite was converged on. The final model order in this simulation was $m_1$, $m_6$, $m_7$, $m_2$, $m_3$, $m_8$, $m_4$ and then $m_5$. Perhaps the AIC objective function is too critical of the model complexity as it seems to always select models according to it.

### 5.1.3 Simulation number 4

In simulation number 4 the PSO algorithm has the adaptive inertia weight but uses the SSE as the objective function.

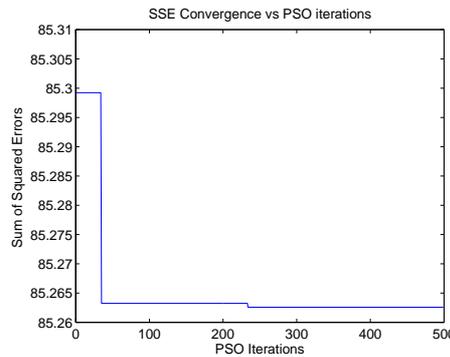

**Figure 8**. Squared sum of Errors (SSE) vs. PSO iterations

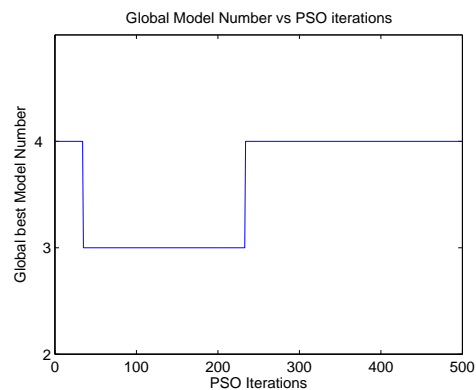

**Figure 9**.SSE Global model convergence vs. PSO iterations

The final model order in this simulation was $m_4$, $m_1$, $m_3$, $m_6$, $m_2$, $m_7$, $m_5$ and then $m_8$. Clearly the SSE objective function is not concerned with the complexity of the finite element model. The best model in this case is one of the most complex. It is not easy to directly compare the objective function magnitudes the algorithms converge to. This would have allowed for better analysis of why different models behave so differently under different objective functions. More conclusive decisions on the choice of objective function in this type of updating methodology can only be made with further experiments on different types of objective functions.

### 6. Conclusions

A particle swarm based method of finite element model updating and selection is presented. The method updates finite element model parameters using a stochastic-population based optimization procedure. Each potentially correct model of a structure is treated as an adaptive particle in the finite element model updating problem space. This space is defined by the number of potentially updatable model parameters. A number of simulations, using different objective function, are performed on eight competing models of a particular structure. The particle swarm based optimization approach to finite element model updating offers the researcher an ability to simultaneously update and select the best model in a given group. This is desirable in the cases where multiple competing model updating models of one structure exist. This paper has also highlighted that the choice of objective function is crucial in obtaining a reasonable best model.

## 7. Acknowledgements

We would like to acknowledge the support of National Research Foundation of the Republic of South Africa.

## 7. References

[1] J.L. Beck, K-V. Yuen. **Model selection using Response Measurements: Bayesian Probabilistic Approach**. Journal of Engineering Mechanics, Vol. 130, pages 192-203, 2004.

[2] J.L. Beck, L.S. Katafygiostis. **Updating Models and their uncertainties. I:Bayesian Statistical framework.** Journal of Engineering Mechanics, Vol. 124, No4, pages 455-461,1998.

[3] C. Bishop. **Pattern Recognition and Machine Learning**. Springer, 2006.

[4] N. Datta. **Finite element Model Updating, Eigenstructure Assignment and Eigenvalue Embedding Techniques for vibrating Systems**. Mechanical Systems and Signal Processing, Vol. 16, pages 83-96, 2002.

[5] M. Dorigo, C. Bhlum. **Ant colony optimization theory: A survey**. Theoretical Computer Science, Vol. 344, pages 243-278, 2005.

[6] H-J Escalante,M. Montes, L-E,Sucar. **Particle Swarm Model Selection**. Journal of Machine Learning Research, Vol. 10 (Feb), pages 405-440, 2009.

[7] D.J. Ewins. **Modal Testing: Theory and Practice**. Publisher Letchworth, Research Studies Press, 1984.

[8] J.R. Fonseca, M.I.Friswell, J.E. Mottershead, A.W. Lees. **Uncertainty identification by the maximum likelihood method**. Journal of Sound and Vibration, Vol. 288, pages 587-599, 2005.

[9] M.I. Friswell, J.E. Mottershead. **Finite element model updating in structural dynamics**. Kluwer Academic Publishers: Boston, 1995.

[10] J. Kennedy, R. Eberhart. **Particle Swarm Optimization**. In Proceedings of the International Conference on Neural Networks, Vol. IV, pages 1942-1948, 1995.IEEE.

[11] J.Kennedy,R. Eberhart. **Swarm Intelligence**. Morgan Kaufmann Publishers, Inc., San Francisco,CA,20001.

[12] Kim G-H, Park Y-S. **An automated parameter selection procedure for finite element model updating and its applications**. Journal of Sound and Vibration. Vol. 309, No. 3-5, pages 778-793, 2008.

[13] M.I. Levin, N.A.J. Lieven. **Dynamic finite element model updating using simulated annealing and genetic algorithms.** Mechanical Systems and Signal Processing, Vol. 12, No. 1, pages 91–120, 1998.

[14] M.Link, M,Friswell. **Working Group 1: Generation of validated structural dynamic models-Results of a benchmark study utilising the GARTEUR SM-AG19 testbed**. Mechanical Systems and Signal Processing, 2003, Vol.17, No. 1, pages 9-20.

[15] Ljung L. **System identification: Theory for the user**. Englewood Cliffs, NJ: Prentice hall, 1987.

[16] C.Mares, B.Dratz, J.E. Mottershead, M.I. Friswell. **Model Updating using Bayesian Estimation**. ISAM 2006, Leuven, Belgium, pages 2607-2616, 18-20 September 2006.

[17] C.Mares, J.E. Mottershead, M.I. Friswell. Selection **and Updating of parameters for the GARTEUR SM-AG19 testbed**. 3[rd] International Conference on Identification in Engineering Systems, Swansea, pages 130-141, April 2002.

[18] T. Marwala. **Finite element model updating using particle swarm optimization**. International Journal of Engineering Simulation, Vol. 6(2), pages 25-30, 2005.

[19] T.Marwala, S.Sibisi. **Finite Element Model Updating using Bayesian Framework and Modal Properties**. Journal of Aircraft, Vol. 42, No. 1, pages 275-278. January-February 2005.

[20] L.Mthembu, T.Marwala, M.I. Friswell, S.Adhikari. **Finite element model updating using a Bayesian Approach**. Submitted Mechanical Systems and Signal Processing, August 2009.

[21] M. Muto, J.L Beck. B**ayesian Updating and Model Class Selection for Hysteretic Structural Models Using Stochastic Simulation**. Journal of Vibration and Control, Vol. 14, No. 1-2, pp 7-34, 2008.

[22] I. Murray , Z. Ghahramani . **A note on the evidence and Bayesian Occam's razor**. Technical Report Gatsby Computational Neuroscience Unit GCNU-TR 2005-003. August 2005

[23] I.J. Myung. The importance of complexity in model selection. Journal of Mathematical Psychology, Vol. 44, pages 190-204, 2000.

[24] Y. Shi,R.C. Eberhart. **Empirical Study of particle swarm optimization**. In Proceedings of the Congress on Evolutionary Computation, pages 1945-1949, Piscataway, NJ, USA, 1999. IEEE.

[25] M.S Voss, X. Feng. Arma **model selection using particle swarm optimization and AIC criterion**. In Proceedings of the 15[th] IFAC Triennial World Congress on Automation Control, Barcelona, Spain, 2002.

[26] Titurus B and Friswell MI. **Regularization in model updating**. International Journal for numerical methods in engineering, Vol. 75, pages 440-478, 2008.